%% file: main.tex
\documentclass[DIV=15, bibliography=totoc]{scrartcl}  
\usepackage[a4paper,margin=2.25cm,bottom=2.5cm]{geometry}

\input{configuration}

\input{frontpage}

\begin{document}

\normalem \maketitle  
\normalfont\fontsize{11}{13}\selectfont

\vspace{-1.5cm} \hrule 

\section*{Abstract}

Full waveform inversion (FWI) is a powerful tool for reconstructing material fields based on sparsely measured data obtained by wave propagation. For specific problems, discretizing the material field with a neural network (NN) improves the robustness and reconstruction quality of the corresponding optimization problem. We call this method NN-based FWI. Starting from an initial guess, the weights of the NN are iteratively updated to fit the simulated wave signals to the sparsely measured data set. For gradient-based optimization, a suitable choice of the initial guess, i.e., a suitable NN weight initialization, is crucial for fast and robust convergence. \\
In this paper, we introduce a novel transfer learning approach to further improve NN-based FWI. This approach leverages supervised pretraining to provide a better NN weight initialization, leading to faster convergence of the subsequent optimization problem. Moreover, the inversions yield physically more meaningful local minima. The network is pretrained to predict the unknown material field using the gradient information from the first iteration of conventional FWI. In our computational experiments on two-dimensional domains, the training data set consists of reference simulations with arbitrarily positioned elliptical voids of different shapes and orientations. We compare the performance of the proposed transfer learning NN-based FWI with three other methods: conventional FWI, NN-based FWI without pretraining and conventional FWI with an initial guess predicted from the pretrained NN. Our results show that transfer learning NN-based FWI outperforms the other methods in terms of convergence speed and reconstruction quality. 

\vspace{0.25cm}
\noindent \textit{Keywords:} 
transfer learning, neural network, deep learning, adjoint optimization, Full waveform inversion
\vspace{0.25cm}

\section{Introduction}
\label{sec:sample1}

Civil infrastructure is often deteriorated due to exposure to loads, temperature fluctuations, natural disasters, and potential faults during construction. Assessing the damage present in the infrastructure requires non-destructive testing (NDT)~\cite{park2000impedance, lynch2007overview, kim2007health}. In recent years, attempts have been made to exploit the potential of full waveform inversion (FWI) in NDT. Originating in seismology~\cite{tarantola1984inversion}, FWI infers material models of the earth's interior by fitting simulated seismic waves to measured data sets from active or passive sources~\cite{fichtner2010full,menke2018geophysical}. The approach is adopted in engineering applications~\cite{seidl2017full, He2021NumericalUF, burchner2023immersed, BKK23} utilizing ultrasonic waves to reveal information about unknown defects such as voids, cracks, and inclusions. Extracting detailed information about damage location and severity using FWI is difficult, as the problem is ill-posed and computationally expensive~\cite{virieux2009overview,virieux2017introduction}.\\

Deep learning-based approaches are investigated to enrich FWI, potentially offering a chance to overcome the challenges mentioned above. There are several methods to incorporate neural networks (NNs) into FWI, the most common being using NNs in a supervised manner. In this context, the NN learns to map time series signals, such as displacements or pressures at sensor locations within the domain, to the corresponding field of material coefficients, often referred to as distribution, which describes wave velocities or density~\cite{yang2019deep, wu2019inversionnet, Wang2020VelocityMB, rao2023quantitative}. Another approach uses NNs to enrich the signals with low-frequency data~\cite{jin2018learn,sun2020extrapolated,yang2022deep,li2022high}. Kleman et al.~\cite{kleman2023full} use an NN to predict the actual velocity distribution by inputting the velocity distribution after the first iteration of conventional FWI. A common challenge with supervised approaches is the need for large training data sets. To address this limitation, generative adversarial networks can be utilized~\cite{araya2019deep}. However, due to the absence of an accuracy measure in the supervised learning approach, the resulting predictions are often unreliable, thereby limiting their practical applicability.\\

Another way to take advantage of deep learning methods in FWI is to use them with conventional FWI~\cite{he2021reparameterized, zhu2022integrating,herrmann2023use, Muller2023}. Herrmann et al.~\cite{herrmann2023use} show that inflating the optimization space by using a massively overparameterized NN to predict the coefficients of a piecewise constant material field can serve as an implicit regularization of the optimization problem. Solving the corresponding problem with the Adam optimizer results in fast convergence to physically reasonable local minima. One reason for this may be the inherent bias of NN to learn the low-frequency content of the data~\cite{rahaman2019spectral}. Therefore, the algorithm converges to an image that has fewer high-frequency artifacts. The gradients used for this case are calculated first by the continuous adjoint method, where the derivative of the cost function with respect to the coefficients of the piecewise constant material field is calculated. Then, the gradient of the cost function with respect to the NN parameters is determined by automatic differentiation (see the gradient computation in \Cref{fig:nonpretrain}). In the sequel, we will refer to this approach of using the NN as a parameterization of the material field as \emph{NN-based FWI}.\\

It is possible to further improve NN-based FWI using transfer learning. Transfer learning is a technique in machine learning where an NN trained on one data set is used for a different but related learning task. By doing so, the NN starts the downstream task with weights already adjusted to recognize relevant features, significantly reducing the training time. Simply put, transfer learning uses the knowledge from previous data to accelerate learning for related tasks. Transfer learning has already been utilized in various fields, such as for the classification of text~\cite{nigam2000text,Qasim2022} and images~\cite{shaha2018transfer,Hussain2018}, machine translation~\cite{Barret2018,Surafel2018} and robotics~\cite{Tang2021,Pereida2018,Makondo2018}. This approach can also be used in conjunction with NN-based Full Waveform Inversion (FWI), where pretraining the NN provides a suitable initial guess for the optimization task. The training data set for this includes wave simulations on samples with arbitrary defects. In a recent study by Muller et al.~\cite{Muller2023}, their NN was pretrained directly on sensor data to predict an appropriate initial velocity distribution. A similar approach is described in~\cite{kollmannsberger2023transfer}.\\

In this paper, although a similar idea is used, a significant improvement in the pretraining of the NN is proposed. Instead of simply using raw sensor data to predict an initial material distribution as in~\cite{kollmannsberger2023transfer}, the gradient computed by the adjoint method with respect to the material coefficients of the piecewise constant representation (obtained during the first iteration with a homogeneous material) is utilized. This simplifies the architecture of the NN because the gradient of the conventional FWI from the first iteration has the same shape as that of the domain. The gradient contains an approximation of the shape of the damage (\Cref{fig:pretraining}), which renders the learning task easier. Furthermore, using pretrained NN removes the dependency of the convergence of the NN on the initial weight initialization. The paper at hand thereby proposes a method for accelerating the NN-based FWI~\cite{herrmann2023use} by pretraining the NN. Using the pretrained NN for FWI is called transfer learning with NN-based FWI. The proposed method is compared with three other approaches: conventional FWI, NN-based FWI without pretraining, and conventional FWI with an initial guess from the pretrained NN.\\

The remainder of the paper is organized as follows: In \cref{sec:background}, the background of conventional and NN-based FWI is discussed, including the proposed transfer learning approach. In~\cref{sec:results}, we demonstrate the superior performance of the transfer learning NN-based FWI with the help of several numerical examples. Finally, \cref{sec:conclusion} concludes the paper's main findings. 

\section{Methodology} 
\label{sec:background}

\subsection{Physical model}

In the sequel, the scalar wave equation is used. It describes the propagation of the scalar wave field $u(\boldsymbol{x},t)$ in space $x$ and time $t$ in a medium whose spatial distribution of the density is given by~$\rho(\boldsymbol{x})$. The spatial distribution of the wave speed is denoted as $c(\boldsymbol{x})$, the external force as $f(\boldsymbol{x},t)$. With this notation, the 2D scalar wave equation reads
\begin{equation}
    \rho(\boldsymbol{x}) \ddot{u}(\boldsymbol{x}, t)-\nabla\cdot(\rho(\boldsymbol{x}) c(\boldsymbol{x})^2 \nabla u(\boldsymbol{x}, t)) = f(\boldsymbol{x},t), \quad \boldsymbol{x}=[x,y] \in \Omega, t \in \mathcal{T}, \label{eq:waveequation}
\end{equation}
where $\Omega$ is the 2D spatial domain and $\mathcal{T} = [0, T_{\text{max}}]$ the time interval of interest. Appendix A of~\cite{herrmann2023use} gives a detailed derivation of this equation. B\"urchner et al.~\cite{burchner2023immersed} show that using mono-parameter FWI to only invert for density leads to better identification of defects with strong contrasts such as voids. Therefore, a dimensionless scaling function $\gamma(\boldsymbol{x})$ is introduced to parameterize the scalar wave equation and the following optimization problem. Assuming $\rho_0$ and $c_0$ correspond to undamaged background material, the unknown material field $\gamma(\boldsymbol{x})$ scales the density, i.e., $\rho(\boldsymbol{x}) = \gamma(\boldsymbol{x})\rho_0$, while the wave speed remains constant, i.e.,  $c(\boldsymbol{x})=c_0$. This leads to the density-scaled version of the scalar wave equation, which is used throughout the paper,

\begin{equation}
    \gamma(\boldsymbol{x})\rho_0\ddot{u}(\boldsymbol{x},t)-\nabla\cdot(\gamma(\boldsymbol{x}) \rho_0 c_0^2 \nabla u(\boldsymbol{x}, t)) = f(\boldsymbol{x},t), \quad \boldsymbol{x}=[x,y] \in \Omega, t \in \mathcal{T} \text{.}
    \label{eq:scaledwave}
\end{equation}

The initial and boundary conditions are

\begin{equation}
    u(\boldsymbol{x}, 0) =\dot{u}(\boldsymbol{x}, 0)=0 \qquad \text{on } \Omega, \label{eq:initialconditions} 
\end{equation}

\begin{equation}
    \frac{\partial u(\boldsymbol{x}, t)}{\partial x} =0  \qquad \text{on } \Gamma_{y}\times\mathcal{T}, 
    \label{eq:neumanninx}
\end{equation}
\begin{equation}
    \frac{\partial u(\boldsymbol{x}, t)}{\partial y} =0  \qquad \text{on } \Gamma_{x}\times\mathcal{T} \text{.} \label{eq:neumanniny} 
\end{equation}

\subsection{Numerical configuration}
\label{sec:problem}

\begin{figure}[htb]
  \centering
  \begin{subfigure}[b]{0.45\textwidth}
      \centering
	\input{fig1a}
   \caption{}
   \label{fig:gamma}
   
  \end{subfigure}
  \hfill
  \begin{subfigure}[b]{0.49\textwidth}
      \includegraphics[width=1\textwidth]{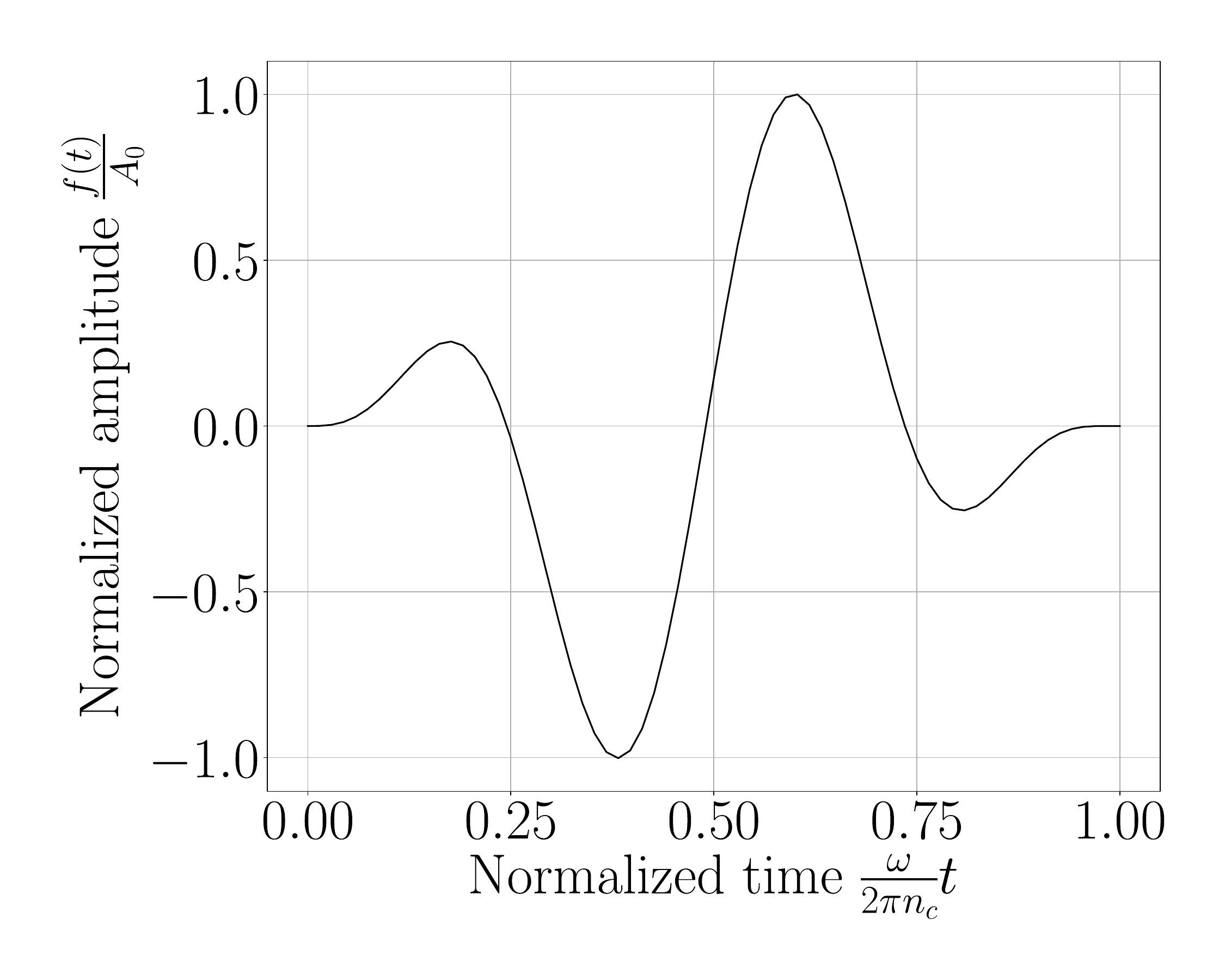}
      \caption{}
   \label{fig:force}
  \end{subfigure}  
  \caption{(a) Set up of the 2D domain. The black dots indicate the position of the sensors in the domain, and the red dots indicate the point of excitation. The blue ellipsoid represents the damage, parameterized using the semi-major and minor axes a,b, the center $(x_c,y_c)$, and the angle $\phi$. (b) Excitation force sine burst with two cycles }
  \label{fig:setup}
\end{figure}

In the paper, all investigations are performed with the same numerical configuration. The setup is shown in~\Cref{fig:setup}.
Considering a 2D domain of $\SI{0.1}{\meter} \times \SI{0.05}{\meter}$, a finite differences (FD) solver is used to solve the corresponding wave equation (as discussed in detail in Appendix A of~\cite{herrmann2023use}). The domain is excited with a two-cycle sine burst at the positions $\boldsymbol{x}^\text{s}$. Introducing the central frequency $k_\text{c}$ with the corresponding angular frequency $\omega = 2 \pi k_\text{c}$ and amplitude $A_0$ (\Cref{fig:force}), the external force reads
\begin{equation}
    f(\boldsymbol{x},\boldsymbol{t}) =
    \begin{cases}
    A_0\sin(\omega t)\sin^2(\frac{\omega t}{2 n_\text{c}}) \delta(\boldsymbol{x} - \boldsymbol{x}^\text{s}) & \text{for} \: 0 \leq t \leq \frac{2 \pi n_\text{c}}{\omega} \\
        0 & \text{for} \: \frac{2 \pi n_\text{c}}{\omega} < t
    \end{cases}
\end{equation}
where $n_\text{c} = 2$ is the number of cycles and $\boldsymbol{x}^\text{s}$ is the source position of the underlying experiment.\\

The undamaged density and wave speed of the material is  $\SI{2700}{\frac{kg}{m^3}}$ and  $\SI{6000}{\,\mathrm{\frac{m}{s}}}$. The central excitation frequency is $\SI{500}{\kilo \hertz}$. The synthetic reference data is generated with simulations on a grid with 512 $\times$ 256 grid points and 2000 time steps with $dt=\SI{3e-8}{\second}$. For the FWI, a different grid configuration has been used to avoid the ``inverse crime''~\cite{wirgin2004inverse}. The domain is discretized with 256 $\times$ 128 grid points, and the time integration is performed with 1000 time steps and a time step length of $dt=\SI{6e-8}{\second}$. The sources and sensors are positioned along the top edge of the domain. There are four sources in total and the simulation with the excitation from one sensor is equivalent to carrying one experiment. The sensors are distributed symmetrically about the center of the plate, 36 grid points apart. 24 sensors spaced 6 grid points apart are placed between the rightmost and leftmost sources. The excitation of one source is equivalent to conducting one experiment. 

\subsection{Conventional FWI}
\label{sec:conv_fwi}
Using conventional FWI, the problem at hand is parameterized by a set of coefficients $\hat{\boldsymbol{\gamma}}$ representing the scaling function $\gamma(\boldsymbol{x})$ at all finite difference grid points. Introducing a misfit functional $\mathcal{L}(\hat{\boldsymbol{\gamma}})$ the corresponding optimization problem is described by
\begin{equation}
    \hat{\boldsymbol{\gamma}}^* = \underset{\hat{\boldsymbol{\gamma}}}{\mathrm{argmin}} \ \mathcal{L} (\hat{\boldsymbol{\gamma}}),
    \label{eq:optprob} 
\end{equation}
where $\hat{\boldsymbol{\gamma}}^*$ is the optimal solution. The misfit functional sums up the squared error at all $N^\text{r}$ receiver positions $\boldsymbol{x}^\text{r}$ for all experiments $N^\text{s}$
\begin{equation}
    \mathcal{L} (\hat{\boldsymbol{\gamma}}; \boldsymbol{u}^{\text{o}}) = \frac{1}{2} \int_{T} \int_{\Omega} \sum_{s=1}^{N_s} \sum_{r=1}^{N_r} \left[ \left(\hat{u}_{\text{s}}(\hat{\boldsymbol{\gamma}};\boldsymbol{x},t)-u_{\text{s}}^{\text{o}}(\boldsymbol{x}^r,t)\right)^2 \delta(\boldsymbol{x}-\boldsymbol{x}^r) \right ]d\Omega dt,
    \label{eq:misfit} 
\end{equation}
where $\hat{u}_{\text{s}}(\hat{\boldsymbol{\gamma}};x,t)$ denotes a single simulated wave field given a set of coefficients $\hat{\boldsymbol{\gamma}}$ and $u_{\text{s}}^{\text{o}}(x^\text{r},t)$ a observed wave field. Measurements across all experiments are summarized by $\boldsymbol{u}^{\text{o}}=\{u_s^\text{o}\}_{s=1}^{N_s}$.

\begin{figure}
  \centering
\includegraphics[width=1\textwidth,trim={0 16cm 0 7cm},clip]{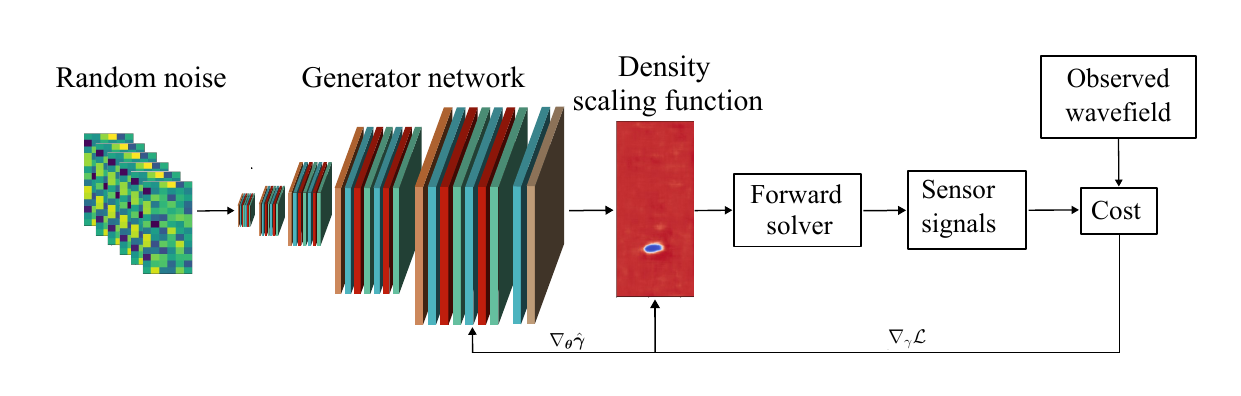} %
  \caption{Workflow of the NN-based FWI. The generator network outputs a density scaling function, which is used to solve the equation using a forward solver. A cost function is calculated based on wavefields from the generator network output and the observed wavefields from the true density scaling function, which is then used to calculate the gradient for backpropagation~\Cref{eq:chainrule}.}
  \label{fig:nonpretrain}
\end{figure}

For the computation of the derivative of the cost function with respect to the material coefficients, the adjoint method is used. For a detailed derivation, see e.g.~\cite{burchner2023immersed}. Given a wave field $u$, the derivative with respect to one coefficient $\hat{\gamma}_i$ is
\begin{equation}
    \frac{d \mathcal{L}}{d \hat{\gamma}_i} = \sum_{s=1}^{N_s} \int_{\mathcal{T}} \left[-\rho_0 \dot{u}_{\text{s,}i}^\dagger \dot{u}_{\text{s,}i} + \rho_0 c_0^2 \nabla u_{\text{s,}i}^\dagger \cdot \nabla u_{\text{s,}i} \right] dt,
\end{equation}
where the subindex $i$ denotes the solution of the corresponding primal and adjoint solution at the grid point $i$. The derivatives at all grid points form the gradient is $\nabla_\gamma \mathcal{L}(\hat{\boldsymbol{\gamma}}; \boldsymbol{u}^\text{o})$ and can readily be used in gradient-based optimization.

\subsection{NN-based FWI}
\label{sec:disc}
\subsubsection{Hybrid gradient computation}
\label{sec:hybrid}
In the NN-based FWI, an NN parameterized by $\boldsymbol{\theta}$ is used to predict the scaling function coefficients, i.e., $\hat{\boldsymbol{\gamma}}(\boldsymbol{\theta})$. To this end, consider an NN $N_\gamma$ that predicts the material coefficients at all grid points: 
\begin{equation}
    \hat{\boldsymbol{\gamma}}(\boldsymbol{\theta}) = N_\gamma(\boldsymbol{\theta};\boldsymbol{k}),
    \label{eq:reprgamma}
\end{equation}
where $\boldsymbol{\theta}$ contains all parameters of the NN and $\boldsymbol{k}$ is the input as shown in~\Cref{fig:decoder}. Therefore, the optimization problem becomes
\begin{equation}
    \boldsymbol{\theta}^* = \underset{\boldsymbol{\theta}}{\mathrm{argmin}} \ \mathcal{L} (\hat{\boldsymbol{\gamma}}(\boldsymbol{\theta}); \boldsymbol{u}^\text{o}).
    \label{eq:optprob} 
\end{equation}
The gradient of the cost function with respect to the NN parameters $\nabla_{\boldsymbol{\theta}}$ can be computed with the chain rule
\begin{equation}
    \nabla_{\boldsymbol{\theta}} \mathcal{L}(\hat{\boldsymbol{\gamma}}; \boldsymbol{u}^\text{o}) = \underbrace{\nabla_{\hat{\boldsymbol{\gamma}}}\mathcal{L}(\hat{\boldsymbol{\gamma}}; \boldsymbol{u}^\text{o})}_{\text{\normalfont adjoint method}} \cdot \underbrace{\nabla_{\boldsymbol{\theta}}\hat{\boldsymbol{\gamma}}}_{\substack{ \textrm{automatic}\\\textrm{differentiation}}} \text{.}
    \label{eq:chainrule}
\end{equation}
As described in detail in~\cite{herrmann2023use}, the first term $\nabla_{\gamma}\mathcal{L}(\hat{\boldsymbol{\gamma}}(\boldsymbol{\theta}); \boldsymbol{u}^\text{o})$ is obtained through the adjoint method, while the latter $\nabla_{\boldsymbol{\theta}}\hat{\boldsymbol{\gamma}}(\boldsymbol{\theta})$ with automatic differentiation. The NN used to predict the density scaling function $\gamma$ (\Cref{eq:reprgamma}) has a generator type architecture~\cite{badrinarayanan2017segnet}. The details of the generator architecture can be found in the Appendix~\Cref{sec:decoder}. The input $\boldsymbol{k}$ to the generator network is a tensor of size $128\times8\times4$ filled with random values sampled from a normal distribution with zero mean and variance one. The output is the spatial distribution of the density scaling function, a tensor of size $1 \times 256 \times 128$ equivalent to the size of the FD grid. The Adam optimizer~\cite{Kingma2014AdamAM} available in the PyTorch library \cite{NEURIPS2019_9015} is used to solve the optimization problem. The NN-based FWI method is summarized in~\Cref{fig:nonpretrain}.

\subsubsection{Transfer learning NN-based FWI}
\label{sec:transfer_fwi}

\begin{figure}[htb]
  \centering
  \fontsize{10pt}{10pt}\selectfont
  \adjustbox{trim=0cm 0cm 0cm 0cm}{%
  \includegraphics[width = 1\textwidth,trim={0 21cm 0 1.8cm},clip]{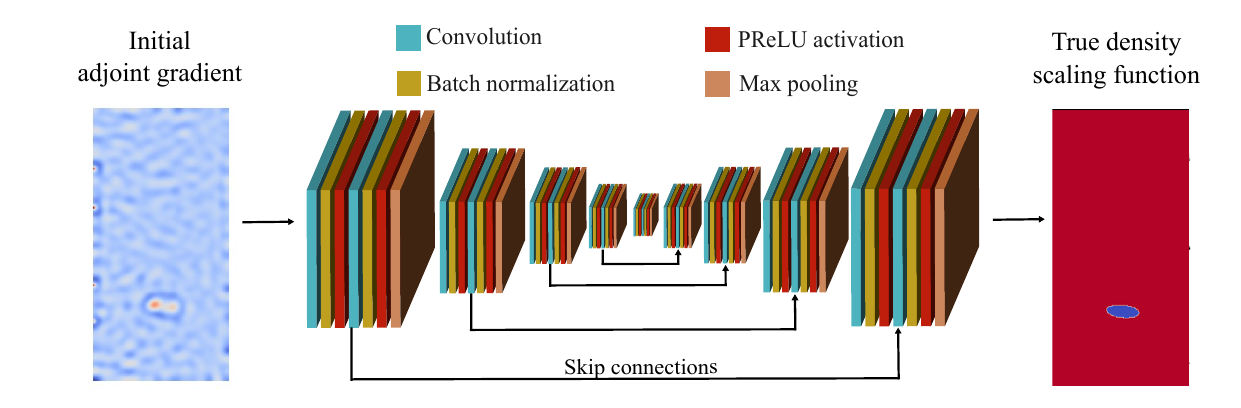}}
  \caption{Pretraining of the U-Net. The supervised training uses the adjoint gradient from the first iteration of the conventional FWI as the input, and the true density scaling function is the output The U-Net is trained on 800 samples.}
  \label{fig:pretraining}
\end{figure}

Transfer learning NN-based FWI uses a pretrained U-Net (\Cref{fig:pretraining}) to parameterize the density scaling function $\gamma$. The knowledge from a training data set is transferred to provide good weight initialization of the NN for the downstream task of FWI.\\

\begin{figure}[htb]
  \centering
  \fontsize{10pt}{10pt}\selectfont
  \adjustbox{trim=0cm 0cm 0cm 0cm}{%
  \includegraphics[width = 1\textwidth,trim={0 16cm 0 8cm},clip]{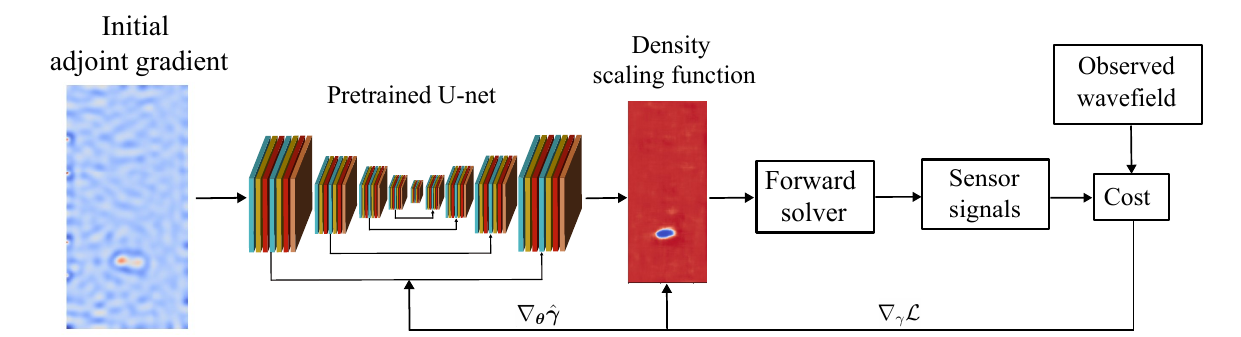}}
  \caption{Workflow of the transfer learning NN-based FWI. The pretrained U-Net is used for the FWI. The weights of the pretrained U-Net are updated over iterations such that the misfit between the observed and simulated wavefields are minimized. }
  \label{fig:workflow_unet}.
\end{figure}

\textbf{Pretraining:} In~\cite{kollmannsberger2023transfer, Muller2023}, the pretraining was carried out using observed wavefields as inputs and the true density distribution as the output. This already improved NN-based FWI. Yet, further improvements are possible by using the adjoint gradient $\nabla_\gamma \mathcal{L}(\hat{\boldsymbol{\gamma}}_0; \boldsymbol{u}^\text{o})$ from the first iteration as the input where the material field $\hat{\boldsymbol{\gamma}}_0$ is equal to 1 throughout the entire domain. This simplifies the design of the NN architecture since the shape of the gradient has the identical dimension as the discretized domain of computation itself, thus allowing the choice of the same shape for input and output (\Cref{fig:workflow_unet}). The gradient also already contains an approximation of the damage. This causes the initial guess to be closer to the sought density scaling function $\boldsymbol{\gamma}^\text{true}$. Therefore, the training data set is $D = \{\nabla_{\gamma}\mathcal{L}(\hat{\boldsymbol{\gamma}}_0; \boldsymbol{u}^\text{o}_i),\boldsymbol{\gamma}^\text{true}_i\}_{i=1}^N$ where $N$ is the number of training samples. The supervised pretraining is carried out using the mean squared error (MSE) over all $N$ samples as loss function given as
 \begin{equation}
     \mathcal{L}_{D} = \frac{1}{N}\sum_{i=1}^{N}\Vert\boldsymbol{\gamma}^\text{true}_i-\hat{\boldsymbol{\gamma}}_i\Vert_2^2,
     \label{eq:pretrainloss}
 \end{equation}
 where $\hat{\boldsymbol{\gamma}}$ is the output of the NN $N_{\gamma}$. For this purpose, the NN, as shown in~\Cref{fig:pretraining}, is used. Note that a different architecture is used in transfer learning NN-based FWI as compared to NN-based FWI, since the input tensor for pretraining is the same size as the output tensor, which was not the case earlier. Supervised training aims to minimize the average loss over the entire training data set. Pretraining the U-Net in such a manner requires one full gradient computation of the first conventional FWI iteration. Therefore, preparing the dataset for pretraining is more expensive than the one carried out in~\cite{kollmannsberger2023transfer} since one adjoint simulation has to be carried out in addition to every forward simulation. However, it leads to better results as demonstrated in~\Cref{sec:results} because it provides a better initialization of the NN. \\

A U-Net-based convolutional NN architecture~\cite{Ronneberger2015} is employed for pretraining. More details about this architecture are in the Appendix (\ref{sec:unet_arch}). The U-Net employs skip connections (\Cref{fig:pretraining}) to address the vanishing gradient problem. As previously mentioned, the input tensor is the adjoint gradient from the first conventional FWI iteration $\mathcal{L}(\hat{\boldsymbol{\gamma}}_0; \boldsymbol{u}^\text{o})$, and the output is the predicted density scaling function $\hat{\boldsymbol{\gamma}}$. The U-Net is trained on a data set containing one ellipsoid-shaped damage parameterized using the semi-major and minor axes a and b, the center $x_c, y_c$, and the rotation $\theta$, see \Cref{fig:gamma}. The mean squared loss function is used along with RMSprop~\cite{Graves2013GeneratingSW} optimization to train the network. The training of the U-Net is carried out with 800 samples for 100 epochs using a batch size of 80 and requires around 240 seconds on a Nvidia Quadro RTX 8000. More details about the optimal number of epochs and samples used for pretraining can be found in Appendix~\ref{app:epochs_pretrain} and~\ref{app:samples_pretrain}, respectively.\\

Once the U-Net is trained, it is used for the downstream task of FWI. The optimization then updates the weights of $N_{\gamma}$ based on the FWI loss function~\Cref{eq:chainrule}. The updated \Cref{eq:reprgamma} is:

 \begin{equation}
    \hat{\boldsymbol{\gamma}}(\boldsymbol{\theta}^p) = N_{\gamma}(\boldsymbol{\theta}^p;\nabla_{\gamma}\mathcal{L}(\hat{\boldsymbol{\gamma}}_0; \boldsymbol{u}^\text{o}))
    \label{eq:pretrained_gamma}.
\end{equation}

In the following, the pretrained U-Net is used to perform NN-based FWI (\Cref{fig:workflow_unet}). The input to the pretrained network is the adjoint gradient from the first conventional FWI iteration. However, the loss function is changed to~\Cref{eq:misfit}, and the gradient used to update $\boldsymbol{\theta}^p$ is used for the NN-based FWI outlined in~\Cref{eq:chainrule}.

\subsubsection{Conventional FWI with initial guess}
\label{sec:conventionalfwi_initguess}
\sloppy For comparison, the initial guess provided by the pretrained network is used with conventional FWI. Therefore, the prediction $\hat{\boldsymbol{\gamma}}$ from the pretrained network $N_{\gamma}(\boldsymbol{\theta}^p;\nabla_{\gamma}\mathcal{L}(\hat{\boldsymbol{\gamma}}_0; \boldsymbol{u}^\text{o}))$ is used as starting model.

\subsubsection{Importance of NN initialization}
\begin{figure}[htb]
  \centering
  \fontsize{10pt}{10pt}\selectfont
  \adjustbox{trim=0cm 0cm 0cm 0cm}{%
  \includegraphics[width = 1\textwidth,trim={3.6cm 19cm 4.6cm 7cm},clip]{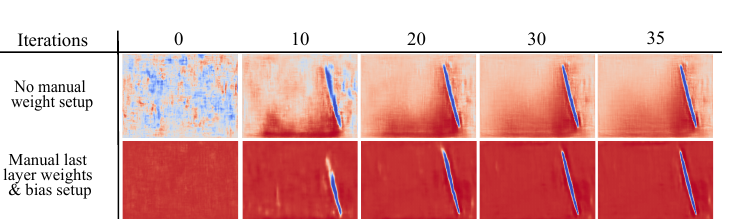}}
  \caption{Effect of the weight initialization on the reconstruction of the density scaling function $\gamma$ over iterations.}
  \label{fig:weightinit}
\end{figure}

Weight initialization for NNs is important to achieve fast convergence and good accuracy. Using the weight initialization suggested by Glorot in~\cite{glorot2010understanding} is common practice. This avoids the vanishing gradient problem but does not form an optimal starting point for FWI, as the random weights lead to the noisy initialization patterns depicted in the upper left part of~\Cref{fig:weightinit}. On the contrary, an expected outcome is that the material remains undamaged almost everywhere except for locally confined flaws. An approximation to undamaged material can be provided by additionally initializing the weights of the last layer using a standard normal distribution with a narrow standard deviation of 0.01 and setting a rather high bias of, e.g. three. This removes most of the noise from the initialization of the density scaling function and provides a better starting point for NN-based FWI. As a last step, a sigmoid function maps the values to the range between zero and one. Furthermore, setting the weight of the last layer causes the method to be less dependent on the initialization of the weights. \Cref{fig:weightinit} shows the effect of the weight initialization on the reconstruction of a damage case within 35 iterations. This weight initialization is used for the generator network for NN-based FWI (\Cref{fig:nonpretrain}) as well as for the U-Net employed for transfer learning NN-based FWI (\Cref{fig:pretraining}). In the case of transfer learning NN-based FWI, the weights of the last layer are initialized in this way during pretraining, which improves the convergence of the training process.

\section{Results and discussion} 
\label{sec:results}

\subsection{Damage cases}

\begin{figure}[htb]
  \centering
  \fontsize{10pt}{10pt}\selectfont
  \adjustbox{trim=0cm 0cm 0cm 0cm}{%
  \includegraphics[width = 1\textwidth]{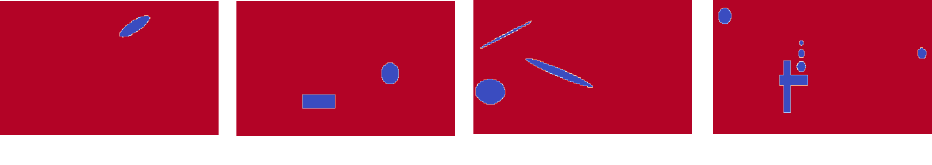}}
  \caption{Four damage cases used for the case study to compare the four methods listed above. The cases are designed in increasing order of complexity.}
  \label{fig:dmg_all}
\end{figure}

To compare all four methods, i.e., conventional FWI (\Cref{sec:conv_fwi}), NN-based FWI (\Cref{sec:hybrid}), conventional FWI with an initial guess from pretraining (\Cref{sec:conventionalfwi_initguess}), and transfer learning NN-based FWI (\Cref{sec:transfer_fwi}), four damage cases are used, see~\Cref{fig:dmg_all}. The aim is to quantify the methods with respect to their reconstruction quality and convergence speed. The red color represents the region where the density scaling function has a value of 1, and the blue color represents a void/damage, i.e., a value of $10^{-5}$ the scaling function. \\

The four case studies are presented to assess the performance of transfer learning NN-based FWI. Then, the four methods are compared quantitatively using an average of over 100 test cases. Apart from the four case studies, we also compare the methods quantitatively with an average performance over 100 test cases. These cases were not part of the training data set used to pre-train the U-Net. All methods were executed on an Nvidia Quadro RTX 8000, and each case required ${\sim}$\SI{100}{seconds} to complete.

\begin{figure}[H]
  \centering
  \fontsize{10pt}{10pt}\selectfont
  \adjustbox{trim=1cm 0cm 0cm 0cm}{%
  \includegraphics[width = 1.0\textwidth,trim={2cm 11cm 5cm 0},clip]{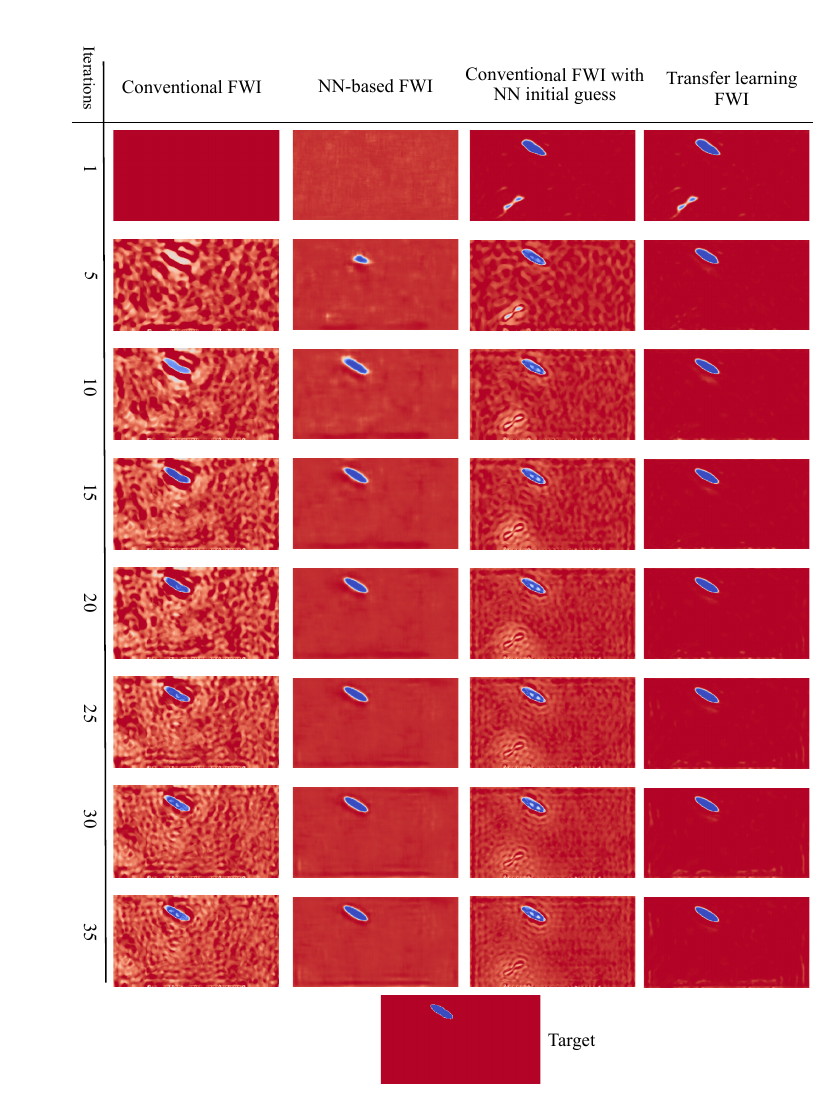}}
  \caption{Case 1: Comparison of the four methods for the shown true density scaling function with one elliptical damage for 35 iterations.}
  \label{fig:case9}
\end{figure}

\subsection{Case 1}
\label{casestd1}
\Cref{fig:case9} compares the four methods for the first damage case. The test case is similar to but not a sample from the data used for the pretraining of the network described in~\Cref{sec:transfer_fwi}. Therefore, it is expected that the pretrained network will perform better than other methods for this case. The first column in~\Cref{fig:case9} shows that conventional FWI recovers the damage within 10-15 iterations. However, the reconstruction suffers from artifacts, which only reduce slightly with increasing iterations. The second column shows the individual snapshots of NN-based FWI without pretraining, which recovers the damage within 10 iterations with good accuracy and without any artifacts in the reconstruction.
The third column shows the results of the corresponding iterations for conventional FWI with an initial guess from the pretrained network. As expected, providing a good initial guess improves the adjoint method's reconstruction and helps recover the damage faster than without pretraining. Finally, the fourth column shows the results from transfer learning NN-based FWI. The reconstruction of the damage is much faster; within 5 iterations, the reconstructed shape of the damage is very accurate, leading to the best results for this case.  

\subsection{Case 2}
\label{casestd2}
\Cref{fig:classic_hybrid_casesquare} shows the comparison of the four methods for a damage case containing a rectangular-shaped hole and a circular hole. Note that the training set does not include a square hole; only elliptical holes were included (\Cref{sec:transfer_fwi}). The first column shows conventional FWI, which recovers the shape approximately within 20-25 iterations. However, there are strong artifacts in the area just above the rectangular damage. The NN-based FWI recovers the rectangular-shaped hole and the circular hole within 20 iterations with good accuracy. The conventional FWI with an initial guess from the pretrained network performs much better than without pretraining and achieves a good reconstruction within 20 iterations with less artifacts in the reconstructed image. Nevertheless, the circular hole is not reconstructed very well. Finally, the transfer learning NN-based FWI performs the best. It achieves an almost perfect reconstruction within 10 iterations without any artifacts. As compared to the NN-based FWI without pretraining, the reconstructed shape is close to the true shape, thus demonstrating that the pretrained network performs very well for cases slightly outside the distribution of data on which it is trained.

\subsection{Case 3}
\label{casestd3}
\begin{figure}[H]
  \centering
  \fontsize{10pt}{10pt}\selectfont
  \adjustbox{trim=1cm 0cm 0cm 0cm}{%
  \includegraphics[width = 1\textwidth,trim={2cm 11cm 5cm 0},clip]{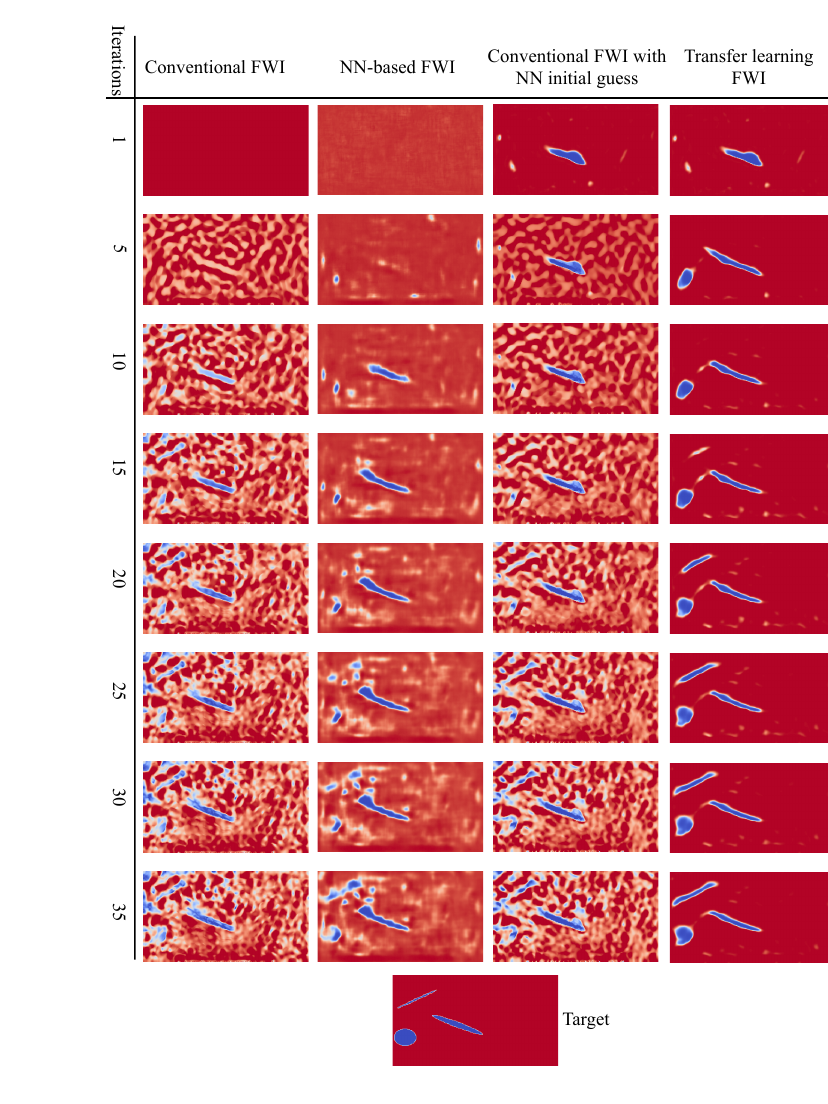}}
  \caption{Case 3: Comparison of the four methods for the shown true density scaling function with 3 elliptical damages for 35 iterations.}
  \label{fig:case15_3dmg}
\end{figure}

For the third case study (\Cref{fig:case15_3dmg}), three holes of different sizes need to be reconstructed. This renders the task to be quite different from the training data. The conventional FWI can roughly recover the rightmost damage while failing to recover the damages on the left side within 35 iterations. There are a lot of artifacts in the reconstruction and it is difficult to infer the correct shape and location of the damages from the reconstruction. The NN-based FWI can recover one damage shape within 15 iterations but struggles to reconstruct the other two. However, in this case, the reconstruction is better than conventional FWI, although the reconstructed shape of the damages is not accurate. The conventional FWI with an initial guess from the pretrained NN, delivers a similar reconstruction of the damage without pretraining. By contrast, the transfer learning NN-based FWI can quickly recover all the locations of the holes with good accuracy within 20 iterations. It improves the reconstructed shape of the damages in the subsequent iterations. The damaged shape is recovered more accurately than other methods, and the artifacts are negligible. Therefore, the transfer learning NN-based FWI performs best among all the methods in this case.

\subsection{Case 4: Is transfer learning NN-based FWI always the best?}
\label{casestd4}
\begin{figure}[H]
  \centering
  \fontsize{10pt}{10pt}\selectfont
  \adjustbox{trim=1cm 0cm 0cm 0cm}{%
  \includegraphics[width = 1\textwidth,trim={2cm 9.5cm 5cm 2cm},clip]{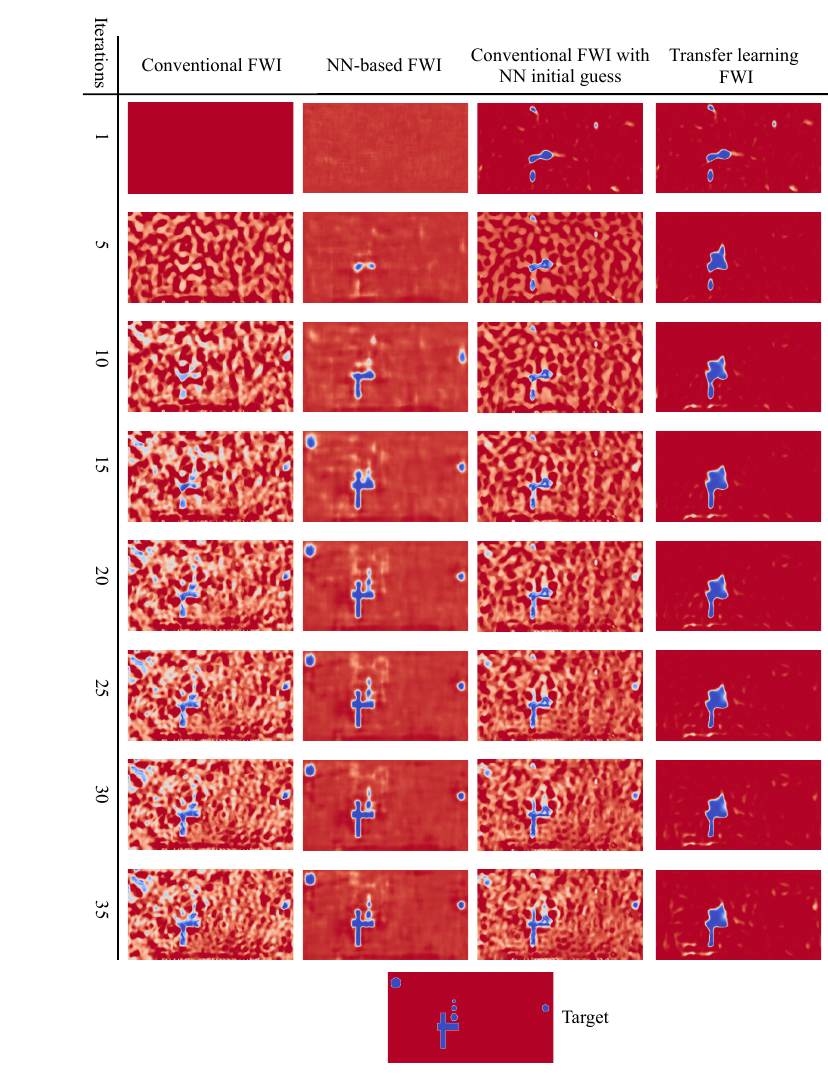}}
  \caption{Case 4: Comparison of all the methods for a manually generated case with multiple damages for 35 iterations.}
  \label{fig:case_diff}
\end{figure}

In very few cases, the transfer learning NN-based FWI does not perform as well as the other methods. The initial guess provided by the network is too inaccurate in some cases where the sought damage is \emph{very} different from those of the training data. One such example is shown in~\Cref{fig:case_diff}. The figure depicts two intersecting rectangular damages with three circular holes with a decreasing radius located on a straight line. Furthermore, there are two additional circular holes on the left and the right sides, resulting in six different damages to be recovered. This is a difficult case for FWI in general, specifically for the pre-trained network, because the training data included holes with only a single damage. Additionally, the smaller circular holes are located in the bigger circular hole's shadow region, making it more challenging to reconstruct the second and third holes.\\ 

The conventional FWI can partially recover the rectangular-shaped damage and one hole on the side. The circular holes behind the rectangular damage are not recovered. The NN-based FWI delivers a fast reconstruction of the whole damaged region using only 25-30 iterations to provide high accuracy. In the case of the conventional FWI with an initial guess from the pretrained network, the initial guess is inaccurate. The recovered damages are partially accurate for the rectangular-shaped damages. The circular holes on each side are also partially recovered. For the transfer learning  NN-based FWI, the method could only recover the shape of the rectangular-shaped damage and missed all the other damages. Upon increasing the total number of iterations, it is observed that it can recover the circular holes on both sides after 55 iterations. Still, it cannot recover the two intersecting rectangular damages with three circular holes behind each other with a decreasing radius ever after 100 iterations. This is one of the few cases where the damage reconstruction is unsatisfactory. However, the NN-based FWI without pretraining performs the best in this case, clearly indicating the advantage of using NN to discretize the material field. 

\begin{figure}[H]
  \centering
  \fontsize{10pt}{10pt}\selectfont
  \adjustbox{trim=1cm 0cm 0cm 0cm}{%
  \includegraphics[width = 1\textwidth,trim={2cm 9.8cm 5cm 0},clip]{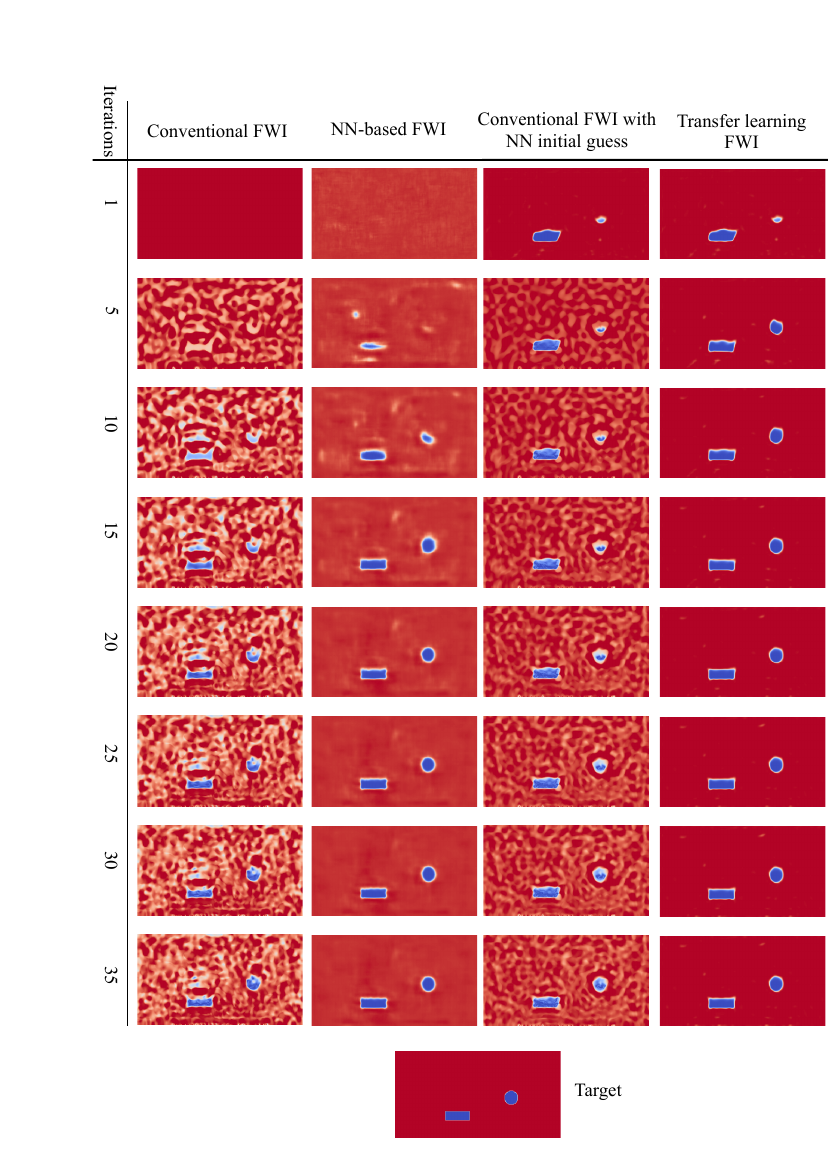}}
  \caption{Case 2: Comparison of the four methods for the shown true density scaling function with a square and a circular damage for 35 iterations.}
  \label{fig:classic_hybrid_casesquare}
\end{figure}

\subsection{Comparison over 100 test cases}
\label{series}

\begin{figure}[htb]
  \centering
  
	\input{comp_cost_mse}
  \caption{The left figure shows the log of the cost for four methods over the iterations averaged over 100 cases containing one elliptical damage (similar to \Cref{casestd1}). Similarly, the right figure shows the log of the mean squared error between the predicted density scaling function $\hat{\boldsymbol{\gamma}}$ and the true density scaling function $\boldsymbol{\gamma}^{\text{true}}$ at each iteration.}
  \label{fig:comp_all_mse_cost}
\end{figure}
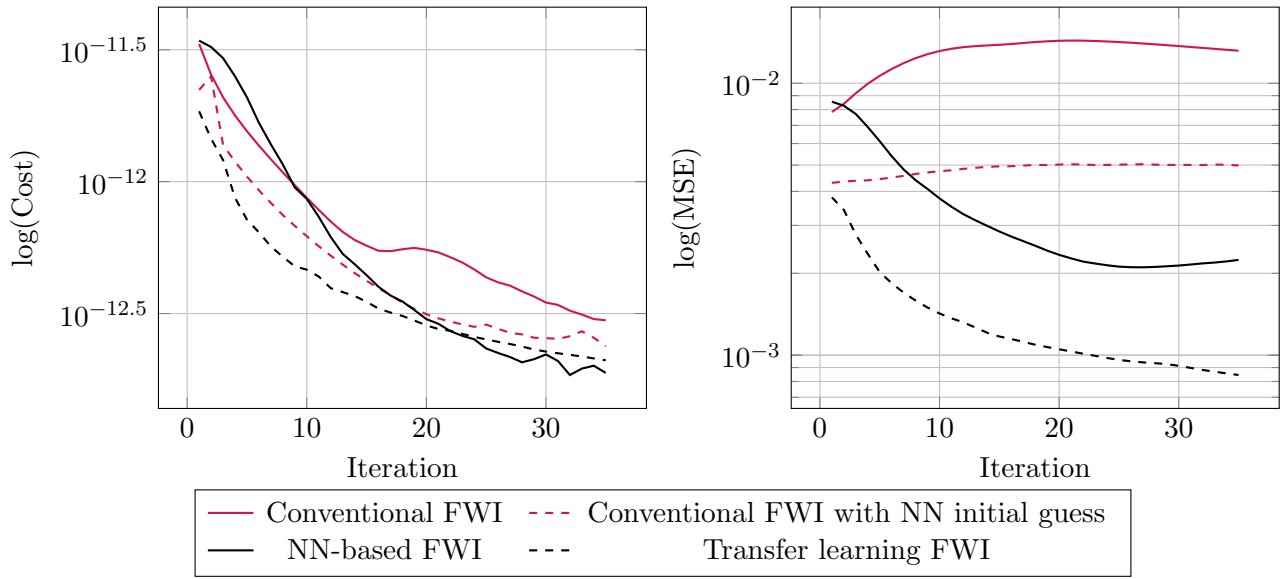

FWI is carried out for 100 test cases to compare the four methods quantitatively, each undergoing 35 iterations. The cost function and the mean squared error between the reconstructed and true density scaling function over the iterations are used as a measure of comparison. Results are depicted in~\Cref{fig:comp_all_mse_cost}.\\

The cost function for the transfer learning NN-based FWI reduces the fastest till 20 iterations. After 35 iterations, the cost is comparable for NN-based FWI and the transfer learning NN-based FWI. However, the mean squared error is the lowest for transfer learning NN-based FWI, followed by NN-based FWI. The mean squared error at the start of the iteration is the same for conventional FWI with an initial guess from the pretrained NN and transfer learning NN-based FWI. However, the MSE for conventional FWI without and with initial guess from pretrained NN increases slightly over the iterations due to the artifacts in the reconstructed density scaling function. Conversely, the MSE consistently decreases for NN-based FWI and transfer learning NN-based FWI, indicating good reconstruction of the density scaling function with fewer artifacts.\\ 

In the examples shown here (as well as in~\cite{herrmann2023use}), the reconstruction of a void using NN-based FWI contains considerably fewer artifacts. This is because NNs contain an inherent spectral bias~\cite{rahaman2019spectral} and prefer to learn the low-frequency content of data. On the contrary, the frequencies of the artifacts in the conventional FWI reconstructions are rather high. Therefore, the reconstruction of the NN-based FWI contains a particular kind of high-frequency filter that surprisingly does not inhibit the formation of sharp changes from material to void and vice versa, and, yet, does not exhibit artifacts. 

\subsection{Can transfer learning FWI recover from bad initial guess?}
\label{sec:badinit}
\begin{figure}[htb]
  \centering
  \fontsize{10pt}{10pt}\selectfont
  \adjustbox{trim=0cm 0cm 0cm 0cm}{%
  \includegraphics[width = 1\textwidth,trim={0.8cm 21cm 3cm 4cm},clip]{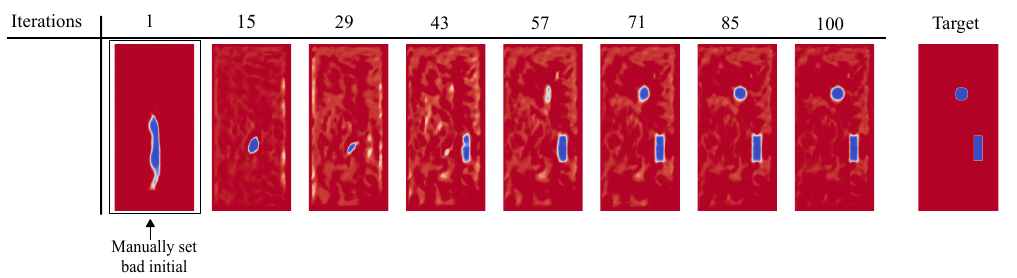}}
  \caption{Performance of the transfer learning FWI for bad initial guess.}
  \label{fig:case_badinit}
\end{figure}

When using a transfer learning framework, a potential issue arises if the pretrained network provides an incorrect initial guess. This can occur when the data given to the pretrained network is different from the training data. To this end, case 2 from \Cref{fig:classic_hybrid_casesquare} is revisited. To emulate a bad pretraining, an incorrect initial guess is set manually for Case 2 (\Cref{fig:dmg_all}) with a subsequent inversion. In Figure \ref{fig:case_badinit}, it is evident that the initial guess in the first iteration differs strongly from the target damage shape. Although the transfer learning FWI required roughly 70 iterations (compared to the 10 iterations with Transfer learning FWI and 25 iterations with NN-based FWI experienced in \Cref{fig:classic_hybrid_casesquare}), it accurately recovers the damage. Similar results were observed for many cases, including Case 1, whereas for some more complicated cases, such as Case 3 and Case 4, the recovered shape was not as good as before. In general, a greater number of iterations is needed to satisfactorily recover the damages when the initial guess is inaccurate. 

\subsection{Drawbacks and limitations}
Transfer learning NN-based FWI also suffers from some drawbacks that are common to any machine learning-based method. It requires extra effort to tune all the hyperparameters involved in the process, such as the learning rate, the learning rate scheduler, gradient clipping, cost scaling, and the number of samples for pretraining. This can be a time-intensive process. \\

As in any other machine learning-based approach, carefully selecting training data is crucial for pretraining the network. Data that is too similar leads to overfitting, with the consequence that the network will not generalize well for out-of-sample data. However, selecting training data can be challenging, particularly if the type of expected damage is unknown. As the domain gets bigger, the task of generating sensible training data for pretraining, as well as hyperparameter tuning, will become more challenging. This indicates potential for further improvement in the architecture of NN models, which are less sensitive to the choice of hyperparameters.

\section{Conclusion}
\label{sec:conclusion}
NN-based FWI~\cite{herrmann2023use} uses an NN to parameterize the density scaling function. It relies on wave signals emitted at sources and received by the sensors to reconstruct flaws. This paper demonstrates that it is possible to improve the NN-based FWI using transfer learning, as introduced in~\cite{kollmannsberger2023transfer}. The transfer learning is achieved with a U-Net, which is pretrained in a supervised manner --- mapping the gradient from the first iteration of conventional FWI to the true density scaling function. The pretraining alleviates the strong dependency of the NN on the initialization of the NN weights. The combination of the NN ansatz in FWI and transfer learning through supervised pretraining greatly improves the convergence and reconstruction quality over conventional FWI. This includes conventional FWI relying on the same initial guess obtained through the pretrained NN. \\

In addition to evaluating the improvements compared to the original transfer-learning-based scheme presented~\cite{kollmannsberger2023transfer}, the transfer learning NN-based FWI is thoroughly compared to conventional FWI both with and without using the same pretrained network to provide a good starting point. In all cases, the presented transfer learning NN-based FWI outperforms conventional FWI. It might be argued that appropriate regularization techniques~\cite{Arridge2019} lead to similar results as the NN-based FWI for conventional FWI. These can, however, equally be applied to both the NN-based FWI and transfer learning FWI. Whether additional regularization boosts the performance of transfer learning FWI and if similar results are obtainable with conventional FWI is the subject of future study.

\section*{Acknowledgements}
The authors gratefully acknowledge the funding provided by the Georg Nemetschek Institut (GNI) through the joint research project DeepMonitor, which finances Divya Shyam Singh and Qing Sun, as well as the Geothermal-Alliance Bavaria (GAB) by the Bavarian State Ministry of Science and the Arts (StMWK), which finances Leon Herrmann. Furthermore, we gratefully acknowledge funds received by the Deutsche Forschungsgemeinschaft under Grant no. KO 4570/1-1 and RA 624/29-1 which support Tim Bürchner. Felix Dietrich would like to acknowledge the funds received by the Deutsche Forschungsgemeinschaft - project no. 468830823.

\section*{Declarations}
\textbf{Conflict of interest} No potential conflict of interest was reported by
the authors.

\section*{Data Availability}
We provide a PyTorch implementation of all methods in \url{https://doi.org/10.5281/zenodo.13150916}.


\renewcommand\thesection{\Alph{section}}
\setcounter{section}{0}
\section{Neural Network Architectures}
\label{sec:sample:appendix}

\subsection{Generator network architecture}
\label{sec:decoder}
\Cref{tab:2dNN} shows the detailed architecture of the generator network used for the NN-based FWI method. The generator network upsamples the input tensor of size $128 \times 8 \times 4$ through its layers to match the finite differences grid of the domain ($1 \times 256  \times 128$). 
\begin{table}[H]
	\caption{2D convolutional NN architecture for the prediction of the indicator $\gamma$. The total number of parameters is $526\,252$.} \label{tab:2dNN}
	\centering
	\begin{tabular}{lll}
		\hline
		\textbf{layer}                          & \textbf{shape after layer}        & \textbf{learnable parameters} \\ \hline
		input                          & $128\times 8\times 4$    & 0                    \\ \hline 
		upsample                       & $128\times 16\times 8$    & 0                    \\ \hline
		2D convolution \& PReLU         & $128\times 16\times 8$    & $147\,584 + 1$          \\ \hline
		2D convolution \& PReLU         & $128\times 16\times 8$    & $147\,584 + 1$          \\ \hline 
		upsample                       & $128\times 32\times 16$  & 0                    \\ \hline
		2D convolution \& PReLU        & $64\times 32\times 16$   & $73\,792 + 1$            \\ \hline
		2D convolution \& PReLU        & $64\times 32\times 16$   & $36\,928 + 1$               \\ \hline 
		upsample                       & $64\times 64\times 32$   & 0                    \\ \hline
		2D convolution \& PReLU        & $64\times 64\times 32$   & $36\,928 + 1$               \\ \hline
		2D convolution \& PReLU        & $64\times 64\times 32$   & $36\,928 + 1$               \\ \hline 
		upsample                       & $64\times 128\times 64$   & 0                    \\ \hline
		2D convolution \& PReLU        & $32\times 128\times 64$   & $18\,464 + 1$              \\ \hline
		2D convolution \& PReLU        & $32\times 128\times 64$   & $9\,248 + 1$               \\ \hline 
		upsample                       & $32\times 256\times 128$ & 0                    \\ \hline
		2D convolution \& PReLU        & $32\times 256\times 128$ & $9\,248 + 1$                \\ \hline
		2D convolution \& PReLU        & $32\times 256\times 128$ & $9\,248 + 1$               \\ \hline 
		2D convolution without padding & $1\times 254\times 126$ & $289 + 1$ \\ 
		\& adaptive Sigmoid & &                   \\ \hline
	\end{tabular}
\end{table}

\subsection{U-Net architecture}
\label{sec:unet_arch}
\begin{table}[h!]
	\caption{2D U-Net convolutional NN architecture used for the pretraining~\cref{fig:pretraining}. The total number of parameters is $784\,039$. $^*$BN: Batch normalization} \label{tab:pretrainNN}
	\centering
	\begin{tabular}{lll}
		\hline
		\textbf{layer}                          & \textbf{shape after layer}        & \textbf{learnable parameters} \\ \hline
		input                          & $1\times 256\times 128$    & 0                    \\ \hline
		2D Convolution \& BN \& PReLU                       & $16\times 256\times 128$    & $160 + 32 + 1$                    \\ \hline
	  2D Convolution \& BN \& PReLU         & $16\times 256\times 128$    & $2320 + 32 + 1$          \\ \hline
		Maxpool 2D    & $16\times 128\times 64$    & $0$          \\ \hline 
  
		2D Convolution \& BN \& PReLU         & $32\times 128\times 64$    & $4640 + 64 + 1$          \\ \hline
		2D Convolution \& BN \& PReLU         & $32\times 128\times 64$    & $9248 + 64 + 1$          \\ \hline
		Maxpool 2D    & $32\times 64\times 32$    & $0$          \\ \hline

        2D Convolution \& BN \& PReLU         & $64\times 64\times 32$    & $18496 + 128 + 1$          \\ \hline
        2D Convolution \& BN \& PReLU         & $64\times 64\times 32$    & $36928 + 128 + 1$          \\ \hline
        Maxpool 2D    & $64\times 32\times 16$    & $0$          \\ \hline

        2D Convolution \& BN \& PReLU         & $128\times 32\times 16$    & $73856 + 256 + 1$          \\ \hline
        2D Convolution \& BN \& PReLU         & $128\times 32\times 16$    & $147584 + 256 + 1$          \\ \hline
        Maxpool 2D    & $128\times 16\times 8$    & $0$          \\ \hline

        2D Convolution \& BN \& PReLU         & $128\times 16\times 8$    & $147584 + 256 + 1$          \\ \hline
        2D Convolution \& BN \& PReLU         & $128\times 16\times 8$    & $147584 + 256 + 1$          \\ \hline
        Upsample    & $128\times 32\times 16$    & $0$          \\ \hline

        2D Convolution \& BN \& PReLU         & $64\times 32\times 16$    & $110656 + 128 + 1$          \\ \hline
        2D Convolution \& PReLU         & $64\times 32\times 16$    & $36928 + 1$          \\ \hline
        Upsample    & $64\times 64\times 32$    & $0$          \\ \hline

        2D Convolution \& BN \& PReLU         & $32\times 64\times 32$    & $27680 + 64 + 1$          \\ \hline
        2D Convolution \& PReLU         & $32\times 64\times 32$    & $9248 + 1$          \\ \hline
        Upsample    & $32\times 128\times 64$    & $0$          \\ \hline

        2D Convolution \& BN \& PReLU         & $16\times 128\times 64$    & $6928 + 32 + 1$          \\ \hline
        2D Convolution \& PReLU       & $16\times 128\times 64$    & $2320 + 1$          \\ \hline
        Upsample    & $16\times 256\times 128$    & $0$          \\ \hline

        2D Convolution \& BN \& PReLU         & $1\times 256\times 128$    & $154 + 2 + 1$          \\ \hline
        2D Convolution \& Sigmoid         & $1\times 256\times 128$    & $10 + 1$          \\ \hline

\end{tabular}
    
\end{table}

Since both the input and the output data are a 2D matrix, a U-Net-based convolutional NN architecture is employed; see \cref{tab:pretrainNN}. The U-Net~\cite{Ronneberger2015} architecture consists of a blocked structure where each block consists of a Convolution layer, a Batch normalization layer, and an activation layer repeated twice, followed by a Max pooling layer. The input and output tensor is $256 \times 128$ in shape. The skip connections (\Cref{fig:pretraining}) are used to overcome the vanishing gradient problem.\\

A small parametric study was carried out to find the best parameters for pretraining the U-Net and the optimum number of samples and epochs required for pretraining. It's worth noting that the optimum performance for the complete method is the performance of the pretrained network for the downstream task, i.e., carrying out the NN-based FWI. After each pretraining task, the network was run for 10 test cases, and the average was compared to the performance of each pretrained network. It was found that the optimum performance was achieved by continuing the pretraining further after the validation error had stopped reducing for the pretraining task. A similar observation was reported in~\cite{neyshabur2020being}.

\subsection{Hyperparameter tuning}
The hyperparameters for each method are tuned separately using 35 iterations to maximize the individual performance (Table \ref{tab:hyp_all}). A learning rate scheduler with a polynomial decay $(\beta\cdot\textrm{epoch}+1)^\alpha$ was also used in this process with $\alpha=-0.5$ and $\beta=0.2$.

\begin{table}[h!]
	\caption{Hyperparameters used for various methods} \label{tab:pretrainNN}
\begin{tabular}{ p{\textwidth/5}p{\textwidth/6}p{\textwidth/6}p{\textwidth/6} p{\textwidth/6} } 
 \hline
Hyperparameter & Transfer learning with NN-based FWI & Conentional FWI with NN initial guess & NN-based FWI & Conventional FWI \\
  \hline
 Learning rate & $5 \times 10^{-4}$ & $5 \times 10^{-2}$ & $ 5 \times 10^{-4}$ & $8 \times 10^{-2}$\\ 
  \hline
 Gradient clipping & $1 \times 10^{-5}$ & $1 \times 10^{-5}$ & $5 \times 10^{-5}$ & $1 \times 10^{-5}$ \\
  \hline
Cost scaling & $1 \times 10^{10}$ & $1 \times 10^{12}$ &$1 \times 10^{8}$ & $1 \times 10^{12}$\\
\hline
\end{tabular}
\label{tab:hyp_all}
\end{table}

\subsubsection{Optimal number of epochs}
\label{app:epochs_pretrain}
To find the optimum number of epochs for pretraining, the network was trained for $50$, $100$, $200$, and $500$ epochs. This was done with three different numbers of samples for pretraining, $50$ samples, $200$ samples, and $500$ samples, to find out if the change in the number of samples is a factor in the optimum number of epochs for pretraining. The hyperparameters used for this parametric study are listed in~\cref{tab:hyp_pretrain}.
\begin{center}
\begin{tabular}{ |c|c| } 
 \hline
Hyperparameter & Value \\
  \hline
 Learning rate & $8 \times 10^{-4}$ \\ 
 Gradient clipping & $5 \times 10^{-5}$ \\
$\alpha$ & -0.5  \\
$\beta$ & 0.2 \\
Batch size & No. of samples / 10 \\
 \hline
\end{tabular}
\captionof{table}{Hyperparameters used for pretraining the U-Net}
\label{tab:hyp_pretrain}
\end{center}

\begin{figure}[htb]
  \centering
  \fontsize{7pt}{7.3pt}\selectfont
  \adjustbox{trim=0cm 0cm 0cm 0cm}{%
  \includegraphics[width =  0.49\textwidth]{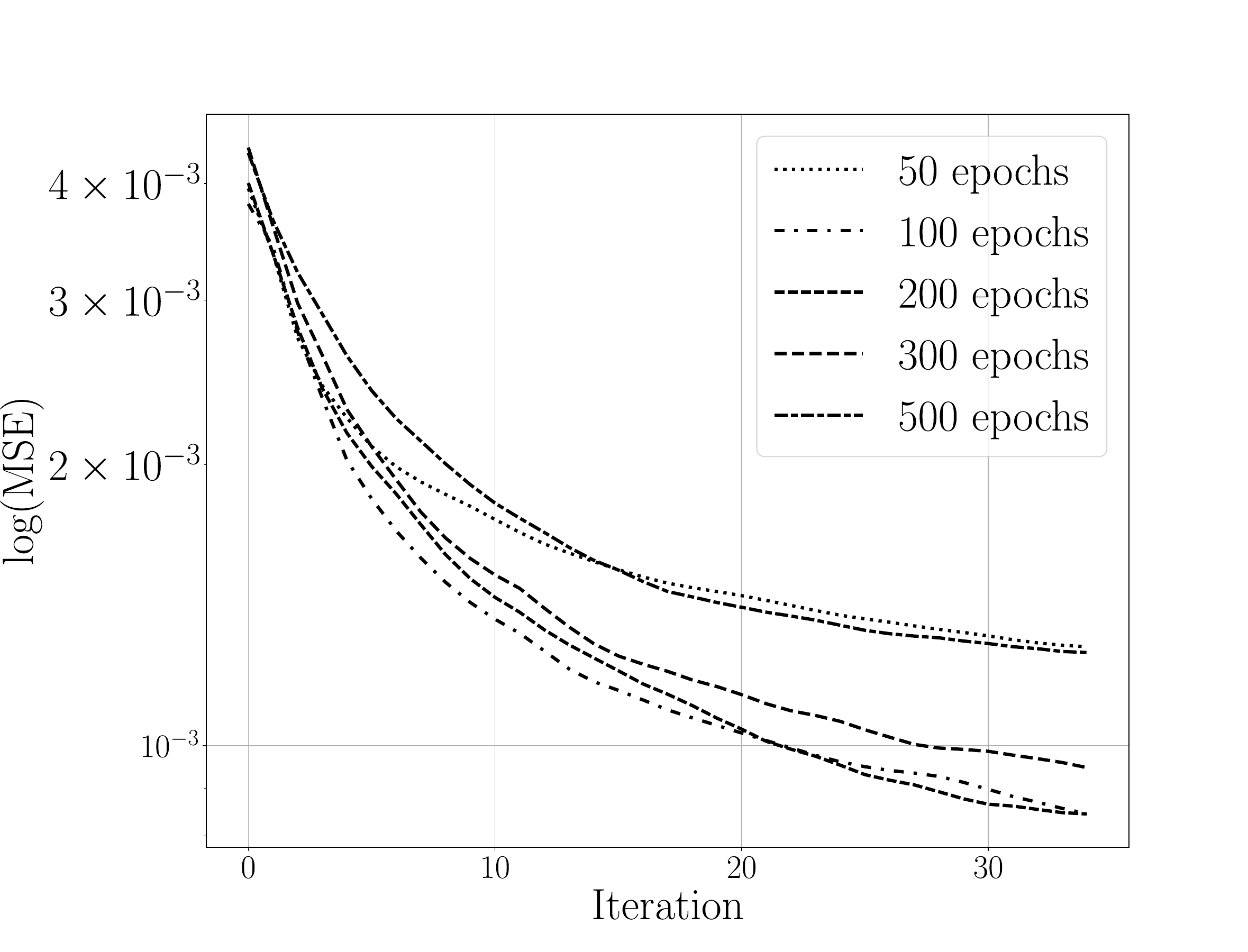}}
  \caption{Performance of the pretrained network with increasing epochs of pretraining with 800 samples}
  \label{fig:pretrain_epochstd}
\end{figure}

\Cref{fig:pretrain_epochstd} shows the result of a parametric study for a U-Net pretrained using $800$ samples and $50,100,200,300,500$ epochs for pretraining averaged over $100$ test cases. The graph also looks similar for $50$ and $200$ samples, with $100$ epochs performing the best for carrying out transfer learning NN-based FWI. Increasing the number of epochs further reduces the accuracy. The initial guess of the network trained for $100$ epochs was the best compared to other cases, and the error reduced much faster than in other cases. Therefore, $100$ epochs have been used to pretrain the U-Net.

\subsubsection{Optimal number of samples} \label{app:samples_pretrain}
\begin{figure}[htb]
  \centering
  \fontsize{7pt}{7.3pt}\selectfont
  \adjustbox{trim=0cm 0cm 0cm 0cm}{%
  \includegraphics[width = 0.49\textwidth]{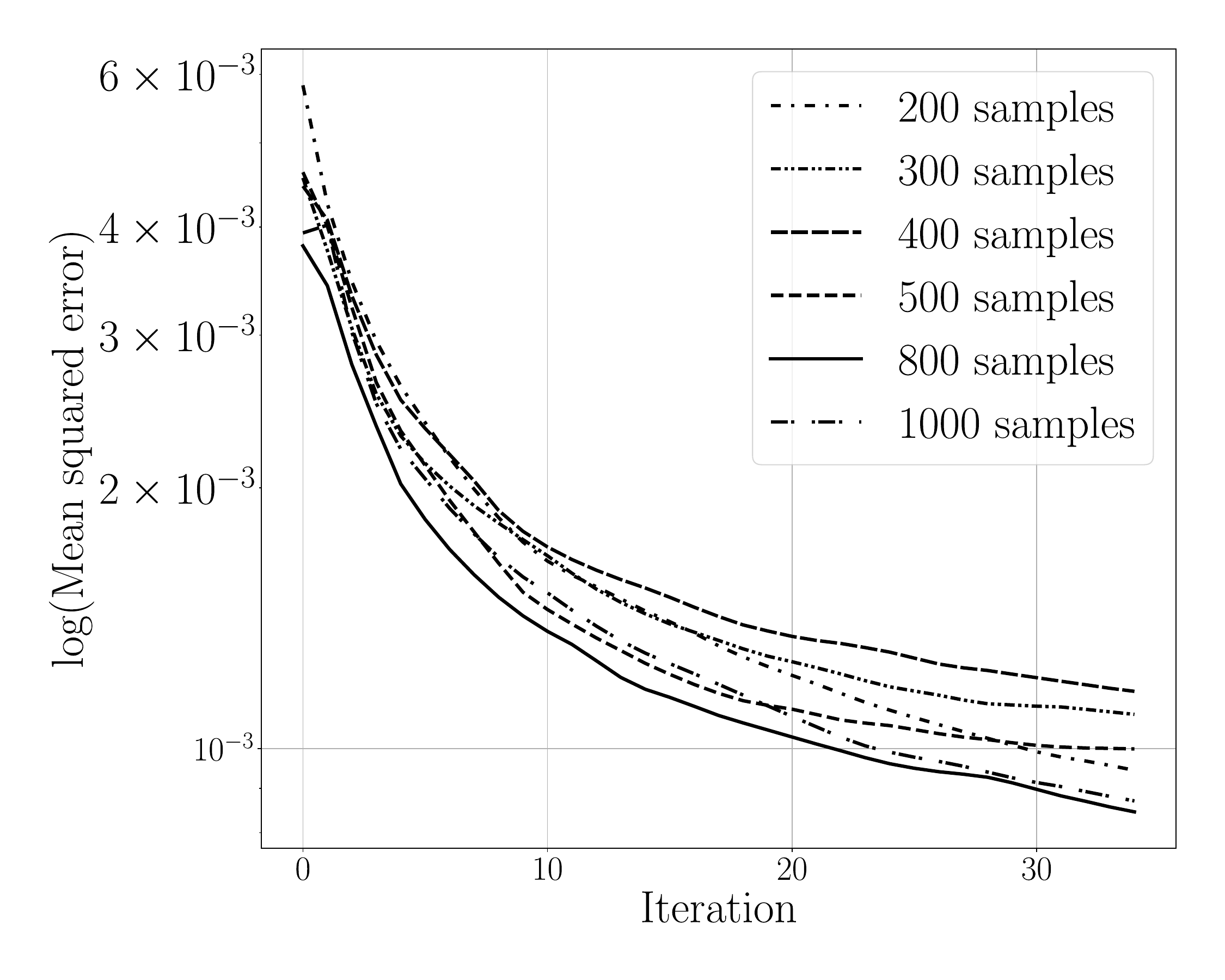}}
  \caption{Performance of the pretrained network with increasing samples for pretraining}
  \label{fig:pretrain_samples}
\end{figure}

A parametric study was conducted to determine the optimal number of data sets used for pretraining the U-Net. This study aims to determine the effect of increasing the number of samples on the performance of the U-Net for NN-based FWI. The number of samples used for pretraining the U-Net is $200$, $300$, $400$, $500$ and $800$. The hyperparameters used for pretraining the network were kept constant for each case. These hyperparameters are shown in~\Cref{tab:hyp_pretrain}. \Cref{fig:pretrain_samples} depicts the mean squared error at each iteration averaged over $100$ test cases. Each curve represents the corresponding amount of samples used for pretraining. \Cref{fig:pretrain_samples} demonstrates the clear pattern that more samples improve the performance of transfer learning NN-based FWI. As can be seen, the MSE error is the least for pretraining using 800 samples for the downstream task. Therefore, the network pretrained from 800 samples is used for final comparison with other methods. 






\end{document}

%% file: configuration.tex
\usepackage[table]{xcolor}
\definecolor{lightblue}{rgb}{0,0.5,1.0}
\definecolor{linkblue}{rgb}{0,0.1,0.6}
\definecolor{citegreen}{rgb}{0,0.4,0.0}
\definecolor{linkred}{rgb}{0.8,0,0.005}
\definecolor{mailviolet}{rgb}{0.3,0,0.35}
\definecolor{tumblue}{rgb}{0,0.396,0.741}
\definecolor{darkgreen}{rgb}{0,0.4,0} 
\definecolor{darkbrown}{rgb}{0.5, 0.396, 0.09}
\usepackage[hyperindex,colorlinks=true,linkcolor=linkblue,citecolor=citegreen,urlcolor=mailviolet,filecolor=linkred]{hyperref}

\usepackage{caption, subcaption}
\usepackage{fancyhdr}
\usepackage{hyperref}
\usepackage[pdftex]{graphicx}

\usepackage[square,comma,nonamebreak,compress,numbers]{natbib}

\usepackage{soulpos}
\usepackage{ulem}
\usepackage{authblk} 
\usepackage{float}

\usepackage{pgfplots}
\usepackage{pgfplotstable}
\usepackage{filecontents}
\usepackage{tikz}
\usepackage{tikzscale}
\usepackage{tikz-3dplot}

\usetikzlibrary{spy}
\usetikzlibrary{intersections}
\usetikzlibrary{arrows,shapes}
\usetikzlibrary{spy}
\usetikzlibrary{backgrounds}  
\usetikzlibrary{decorations}
\usetikzlibrary{decorations.markings}
\usetikzlibrary{positioning}
\usetikzlibrary{patterns}
\usetikzlibrary{calc} 
\usetikzlibrary{quotes} 
\usetikzlibrary{external} 
\usetikzlibrary{matrix}

\pgfplotscreateplotcyclelist{customCycleList}
{%
  {blue, mark=o},
  {red, mark=square},
  {darkgreen, mark=diamond},
  {cyan, mark=diamond},
  {darkbrown, mark=triangle*},
  {black, mark=pentagon},
}

\pgfplotsset{every axis/.append style= {
    cycle list name=customCycleList,
}}

\usepackage{amssymb}
\usepackage{subfiles} 
\usepackage{amsmath}
\usepackage{subcaption}
\usepackage{mathtools}
\usepackage{tablefootnote}
\usepackage{svg} 
\usepackage{adjustbox}
\usepackage{tabularx}
\usepackage{caption}
\usepackage{svg}
\usepackage[symbol]{footmisc}
\usepackage{hyperref}
\usepackage[noabbrev]{cleveref}
\usepackage{float}
\usepackage{tikz}
\usepackage{tikzscale}
\usepackage{tikz-3dplot}
\usepackage{multicol}
\usepackage{siunitx}
\usepackage{xurl}
\usepackage{breakcites}
\usepackage{hyperref}
\usetikzlibrary{spy}
\usetikzlibrary{intersections}
\usetikzlibrary{arrows,shapes}
\usetikzlibrary{spy}
\usetikzlibrary{backgrounds}  
\usetikzlibrary{decorations}
\usetikzlibrary{decorations.markings}
\usetikzlibrary{positioning}
\usetikzlibrary{patterns}
\usetikzlibrary{calc} 
\usetikzlibrary{quotes} 
\usetikzlibrary{external} 
\usetikzlibrary{matrix}
\usepackage{pgfplots}
\usepgfplotslibrary{groupplots}
\pgfplotsset{compat=1.18}

%% file: frontpage.tex
\title{Accelerating Full Waveform Inversion By Transfer Learning}

\author[1]{Divya Shyam Singh\thanks{\href{mailto:leon.herrmann@tum.de}{\texttt{divya.singh@tum.de}},
    Corresponding author}}
\author[1]{Leon Herrmann}
\author[2]{Qing Sun}
\author[1]{Tim Bürchner}
\author[2]{Felix Dietrich}
\author[3]{Stefan Kollmannsberger}

\affil[1]{Chair of Computational Modeling and Simulation, Technical University of Munich, School of Engineering and Design, Arcisstraße 21, Munich, 80\,333, Germany}

\affil[2]{Chair of Scientific Computing in Computer Science (SCCS), School of Computation, Information and Technology, Technical University of Munich, Boltzmannstrasse 3, Garching, 85748, Germany}

\affil[3]{Chair of Data Science in Engineering, Bauhaus-Universit\"at Weimar, Coudraystraße 13 b, 99423, Weimar, Germany}

\newcommand{\publicationDate}{\today}

\date{}
\fancypagestyle{plain}{%
\fancyhf{}%
\fancyfoot[L]{
\vspace{-0.0cm} 
\footnotesize{Preprint submitted to Computational Mechanics.  \hfill 
\publicationDate
\\
}} }

%% file: fig1a.tex
\begin{tikzpicture}
	\def\xc{5}
	\def\yc{1.5}
    \def\a{1.2} 
    \def\b{0.5} 
    \def\angle{45} 
  
	\fill [lightgray, line width=0.3mm] (0,0) rectangle (6,4.2);
    \fill [line width=0.3mm, white, rotate around={\angle:({\xc},{\yc})}] ({\xc},{\yc}) ellipse ({\a} and {\b});
    
    \draw [line width=0.4mm, rotate around={\angle:(\xc, \yc)}] (\xc, \yc) -- (\xc+\a, \yc);
    \draw [line width=0.4mm, rotate around={\angle:(\xc, \yc)}] (\xc, \yc) -- (\xc, \yc+\b);
    \draw [line width=0.3mm, dashed] (\xc, \yc) -- (\xc+0.6, \yc);
    \draw [line width=0.3mm] (\xc+0.6, \yc) arc [radius=0.6, start angle=0, end angle=\angle];
    
    \node [rotate=\angle] at (\xc+0.2, \yc+0.4) {\footnotesize $a$};
    \node [rotate=\angle] at (\xc-0.3, \yc+0.1) {\footnotesize $b$};    
    \node at (\xc+0.4, \yc+0.17) {\footnotesize $\phi$};
    \node at (\xc-0.1, \yc-0.2) {\footnotesize $(x_c, y_c)$};
    

	\node [left] at (\xc-0.2,\yc-0.7) {$\Omega_V$};
	
	\begin{scope}
	\clip (0,0) rectangle (6,4.2);
	\foreach {\r} in {0.3,0.5,...,1.5} {
		\draw [line width=0.2mm, red] (3,4.2) circle (\r cm);
	}
	\end{scope}

	\draw [gray, line width=0.5mm,<->] (0,1) -- (0,0) -- (1,0);
	\node [gray] at (-0.3,0.9) {$y$};
	\node [gray] at (0.9, -0.3) {$x$};

	\fill [red] (3,4.2) circle (0.2cm);

	\foreach {\x} in {0.6,1.2,...,5.4} {
		\foreach {\y} in {4.2} {
			\fill [black] (\x,\y) circle (0.1cm);
		}
	}

	\node at (-0.5,2.1) {$\Gamma_{y_0}$};
	\node at (6.5,2.1) {$\Gamma_{y_1}$};
	\node at (3,-0.3) {$\Gamma_{x_0}$};
	\node at (3,4.6) {$\Gamma_{x_s}$};
	\node at (1,1) {$\Omega$};
	\end{tikzpicture}

%% file: comp_cost_mse.tex
\definecolor{alizarin}{rgb}{0.82, 0.1, 0.26}
\begin{tabular}{rl}
\begin{tikzpicture}[baseline,trim left={1pt}]
    \begin{axis}[width=0.485\columnwidth,
        grid=both,xlabel={Iteration}, ylabel={log(Cost)}, ymode=log]
    \addplot[color=alizarin, thick] file[]  {Images/data_tikz/cost_adjoint.dat}; \label{Conventional FWI}
    \addplot[color=alizarin, thick, dashed] file[] {Images/data_tikz/cost_pretrained_adjoint.dat};\label{Conventional FWI with NN initial guess}
    \addplot[color=black, thick]  file[]  {Images/data_tikz/cost_nontrained.dat}; \label{NN-based FWI}
    \addplot[color=black, thick, dashed] file[] {Images/data_tikz/cost_pretrained_hybrid_800.dat}; \label{Transfer learning FWI}
    \end{axis}
\end{tikzpicture}
    &
\begin{tikzpicture}[baseline,trim left={1pt}]
    \hspace{43pt}
    \begin{axis}[width=0.485\columnwidth,
        grid=both,xlabel={Iteration}, ylabel={log(MSE)}, ymode=log]
    \addplot[color=alizarin, thick] file[] {Images/data_tikz/mse_adjoint.dat}; \label{Conventional FWI}
    \addplot[color=alizarin, thick, dashed] file[] {Images/data_tikz/mse_pretrained_adjoint.dat}; \label{Conventional FWI with NN initial guess}
    \addplot[color=black, thick] file[] {Images/data_tikz/mse_nontrained.dat};  \label{NN-based FWI}
    \addplot[color=black, thick, dashed] file[]{Images/data_tikz/mse_pretrained_hybrid_800.dat}; \label{Transfer learning FWI}
    \end{axis}
\end{tikzpicture}
\\
\end{tabular}

\begin{tikzpicture}
    
\matrix[
      matrix of nodes,
      anchor=north,
      draw,
      inner sep=0.2em,
    ]
    { \ref{Conventional FWI}& Conventional FWI&[5pt]
      \ref{Conventional FWI with NN initial guess}& Conventional FWI with NN initial guess&[5pt]\\
      \ref{NN-based FWI}& NN-based FWI&[5pt]
      \ref{Transfer learning FWI}& Transfer learning FWI&[5pt]\\};
\end{tikzpicture}